\documentclass{article}

\usepackage[preprint]{neurips_2026}

\usepackage[utf8]{inputenc}
\usepackage[T1]{fontenc}
\usepackage{hyperref}
\usepackage{url}
\usepackage{booktabs}
\usepackage{amsfonts}
\usepackage{nicefrac}
\usepackage{microtype}
\usepackage[table]{xcolor}
\usepackage{graphicx}
\usepackage{textcomp}
\usepackage{colortbl}
\usepackage{enumitem}
\usepackage{algorithm}
\usepackage{algpseudocode}
\usepackage{amsmath,amssymb}
\usepackage{multirow}
\usepackage{makecell}
\usepackage{tabularx}
\usepackage[most]{tcolorbox}
\usepackage{upquote}
\tcbuselibrary{skins,breakable,listings}

\definecolor{oursRow}{HTML}{E8F5E9}
\definecolor{ablRow}{HTML}{FFF3E0}
\definecolor{headerRow}{gray}{0.95}
\definecolor{closedRow}{gray}{0.92}
\definecolor{pendingRow}{HTML}{F7F7F7}
\definecolor{ForestGreen}{HTML}{2E7D32}
\definecolor{MutedOrange}{HTML}{EF6C00}

\tcbset{
  guiguardprompt/.style={
    enhanced jigsaw,
    breakable,
    colback=orange!7!white,
    colbacklower=orange!7!white,
    colframe=orange!75!black,
    colbacktitle=orange!80!black,
    coltitle=white,
    interior style={fill=orange!7!white},
    frame style={draw=orange!75!black},
    fonttitle=\bfseries,
    title=title,
    boxrule=0.8pt,
    arc=2pt,
    left=8pt,
    right=8pt,
    top=7pt,
    bottom=7pt
  }
}

\title{Harnessing Pre-Resolution Signals for \\ Future Prediction Agents}

\author{%
\parbox{0.90\textwidth}{\centering
\textbf{Chuyang Wei}$^{1,2}$, \textbf{Maohang Gao}$^{1,2}$, \textbf{Zhixin Han}$^{2}$, \textbf{Kefei Chen}$^{2,3}$, \textbf{Yu Zhuang}$^{2}$\\
\textbf{Minghang Zhu}$^{2}$, \textbf{Haoxiang Guan}$^{1,2}$, \textbf{Yanzhi Zhang}$^{2}$, \textbf{Yilin Cheng}$^{2}$, \textbf{Xiren Zhou}$^{1}$\\
\textbf{Huanhuan Chen}$^{1}$, \textbf{Jian Li}$^{3}$, \textbf{Jiyan He}$^{2}$\thanks{Corresponding authors.},
\textbf{Yu Shi}$^{2}$\footnotemark[1],
\textbf{Yitong Duan}$^{2}$\footnotemark[1],
\textbf{Shuxin Zheng}$^{2}$\footnotemark[1]\\[0.6ex]
$^{1}$University of Science and Technology of China\\
$^{2}$Zhongguancun Academy, Beijing, China \quad
$^{3}$IIIS, Tsinghua University\\[0.5ex]
{\small\texttt{weichy2023@mail.ustc.edu.cn, sz@bza.edu.cn}}
}%
}

\begin{document}

\maketitle

\begin{abstract}
Many high-stakes decisions depend on forecasts made before outcomes are known. In this \emph{future prediction} setting, the central challenge is that public evidence evolves over time, while the main supervision signal arrives only after resolution: the realized outcome mainly assesses final correctness, offering only coarse guidance on what to track, what to verify, and which judgments to leave uncertain along the way. Our key observation is that revisiting the same unresolved question over time creates informative temporal contrasts across evolving evidence and repeated forecasts, exposing what earlier attempts missed before resolution and yielding a diagnostic signal we call the \emph{pre-resolution signal}. We instantiate this idea in Milkyway, a future prediction agent with a persistent \emph{future prediction harness}, an editable external state that stores reusable procedural guidance across revisits to the same unresolved question. As the same unresolved question is revisited, Milkyway extracts pre-resolution signals from evolving evidence and repeated forecasts, uses them to update the harness, and improves later forecasts on that question before resolution. After resolution, the realized outcome serves as a post-resolution check of provisional updates. On the \textsc{FutureX} and \textsc{FutureWorld} benchmarks, Milkyway achieves strong performance against competitive baselines, and a mechanism study suggests that the gains stem from harness evolution driven by pre-resolution signals rather than repeated prediction alone.
\end{abstract}

\section{Introduction}
\label{sec:intro}
Accurately predicting future outcomes matters for many real-world decisions. Governments may need to anticipate economic, public-health, or geopolitical developments, while companies may need to estimate future demand before making production, pricing, or marketing decisions. As language models are increasingly deployed as tool-using agents for open-world information seeking and decision support, a natural question is whether they can also help with questions whose answers are not yet known. We study this setting as \emph{future prediction}: given an unresolved question, an agent must issue a forecast using only time-stamped public information available at prediction time. This differs from many fixed-answer agent benchmarks, where the target answer is already determined at inference time and can often be retrieved or verified directly~\citep{mialon2023gaia,wei2025browsecomp}. In future prediction, by contrast, the relevant evidence is partial and evolving. Useful signals may emerge gradually, while decisive information may still be unavailable when the prediction is made. Even when a question has clear resolution criteria, supervision is typically delayed and coarse: an outcome label appears only after resolution, mainly indicating whether the final forecast was correct, not what the agent should have tracked, verified, or kept uncertain along the way.

Recent work has begun to establish future prediction as a distinct research setting, with progress in both evaluation and learning. On the evaluation side, benchmarks such as FutureX, FutureWorld, and Prophet Arena assess systems on real-world questions whose outcomes are unresolved when forecasts are made and whose public evidence may continue to evolve before resolution~\citep{zeng2025futurex, FutureWorld2026han,prophetarena2025livebenchmark}. On the learning side, recent methods such as Outcome-based Reinforcement Learning and Future-as-Label improve forecasting systems by using realized outcomes after questions resolve~\citep{turtel2025outcome,turtel2026futureaslabel}. Despite this progress, the adaptation signals used by these methods largely arrive only after resolution. Realized outcomes are valuable for judging final correctness, but they provide weak process-level credit assignment: they reveal little about what the agent should have tracked earlier, which evidence it should have sought or verified, and where uncertainty should have been maintained. This motivates looking for additional process-level signals that become available before resolution, as the question and its evidence evolve over time.

\begin{figure*}[t]
    \centering
    \includegraphics[width=\textwidth]{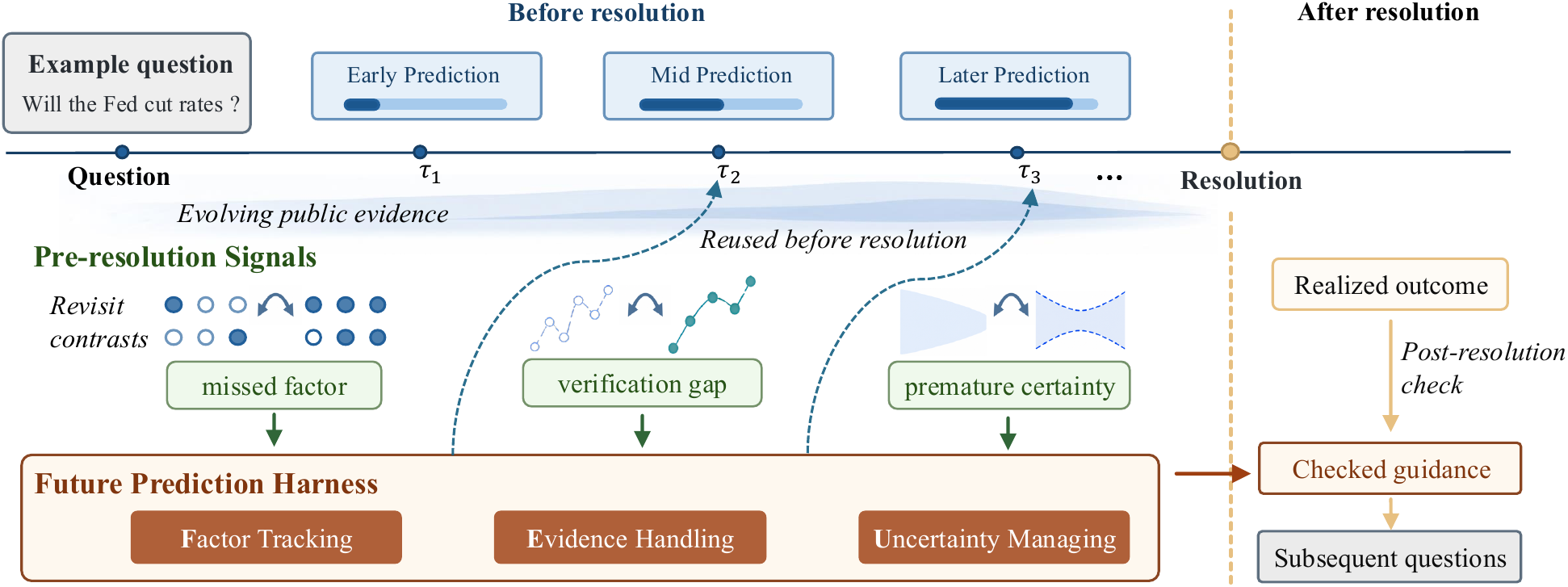}
\caption{\textbf{Pre-resolution signals from revisits in future prediction.}
For the same unresolved question, public evidence evolves while the outcome remains unknown, and contrasts across revisits can reveal missed factors, missed verification, and premature certainty before resolution. \emph{Milkyway} writes these signals into a persistent future prediction harness to guide factor tracking, evidence handling, and uncertainty management at later revisits. After resolution, the realized outcome provides a complementary check on which provisional guidance to retain, revise, or discard.}
\label{fig:milkyway_overview}
\end{figure*}

One natural source of such signal arises when the same unresolved question, under a fixed resolution criterion, is revisited multiple times before it resolves. As public information accumulates, later revisits can expose diagnostic temporal contrasts. Newly available evidence may invalidate earlier assumptions; information that was already available may reveal missed retrieval or verification opportunities; and the continued absence of decisive evidence may indicate where confidence should have remained bounded. Our key observation is that such temporal contrasts across evolving evidence, repeated forecasts, and their supporting rationales can yield informative \emph{pre-resolution signals}. Unlike supervision from realized outcomes, these signals become available while the question remains unresolved. They can therefore inform later revisits to the same question while it remains open, rather than waiting for outcome labels that arrive only after resolution.

We instantiate this idea in \emph{Milkyway}, a future prediction agent equipped with a persistent \emph{future prediction harness}: a compact, editable set of procedural guidance for factor tracking, evidence handling, and uncertainty management across revisits. As the same unresolved question is revisited over time, \emph{Milkyway} compares evolving evidence and forecasts, extracts pre-resolution signals from their temporal contrasts, and applies bounded updates to the harness. These updates can be reused on later revisits to the same question before resolution. \emph{Milkyway} thus separates local diagnosis at revisits from persistent adaptation across revisits. After the question resolves, the realized outcome provides a complementary audit that helps retain, revise, or discard provisional updates. In this way, \emph{Milkyway} shows how future prediction agents can adapt not only from realized outcomes after resolution, but also from signals uncovered while the question is still open.

Our contributions are threefold:
\begin{enumerate}
\item We identify \emph{pre-resolution signals} as a source of process-level adaptation signal for future prediction. These signals arise before an outcome is known, from temporal contrasts across revisits to the same unresolved question under evolving public evidence.
\item We introduce \emph{Milkyway}, a future prediction agent that converts these signals into bounded patch updates to a persistent \emph{future prediction harness}. The harness stores procedural guidance for factor tracking, evidence handling, and uncertainty management across revisits, while post-resolution outcomes provide complementary audits of provisional updates.
\item We evaluate \emph{Milkyway} on the future-prediction benchmarks \textsc{FutureX} and \textsc{FutureWorld}, where it achieves strong forecasting performance against competitive baselines; controlled ablations further show that pre-resolution harness evolution provides value beyond repeated forecasting with the same agent scaffold.
\end{enumerate}

\section{Related work}
\label{sec:related}

\paragraph{Future prediction.}
\emph{Future prediction} asks an agent to answer an unresolved question using only time-stamped public information available at prediction time. Early benchmarks such as ForecastQA and AutoCast formulate forecasting from dated corpora or historical questions~\citep{jin2021forecastqa,zou2022autocast}. Subsequent benchmarks emphasize complementary evaluation desiderata: ForecastBench studies dynamic evaluation with regularly refreshed future-event questions, Bench to the Future provides a hermetic pastcasting setting over large offline corpora, and PROPHET focuses on inferability under retrieved news evidence~\citep{karger2025forecastbench,futuresearch2025btf,tao2025prophet}. More recent benchmarks move closer to live evaluation on real-world questions whose outcomes are unresolved when forecasts are made and whose public evidence may continue to evolve before resolution, including FutureX, FutureWorld, and Prophet Arena~\citep{zeng2025futurex,FutureWorld2026han,prophetarena2025livebenchmark}. Together, these benchmarks help establish future prediction as a distinct setting in which decisive evidence may still be unavailable at prediction time and supervision often arrives only after resolution.

\paragraph{Learning from realized outcomes after resolution.}
A related line of work improves forecasting systems using supervision that becomes available after questions resolve. \citet{halawi2024forecasting} retrieve and aggregate evidence with an LLM and fine-tune on resolved questions to approach human-level accuracy. More recent methods make realized outcomes the explicit learning signal: Outcome-based Reinforcement Learning adapts RLVR to future-event forecasting with noisy delayed rewards~\citep{turtel2025outcome}, and Future-as-Label treats the passage of time as supervision~\citep{turtel2026futureaslabel}. Related efforts likewise leverage resolved questions or realized outcomes to improve forecasting systems after resolution~\citep{chandak2025scaling,scott2026forecasting,unipat2026echo}. These methods highlight the value of post-resolution supervision. Our work is complementary: it studies an additional source of adaptation signal that becomes available before resolution, extracted from temporal contrasts across repeated revisits to the same unresolved question.

\paragraph{Self-improving agents and intermediate feedback.}
A broader line of work improves agents without weight updates by revising editable text state such as reflections, prompts, memories, and skill libraries. Reflexion stores verbal reflections in episodic memory, ExpeL distills reusable lessons, Voyager builds a reusable skill library, and GEPA evolves prompts through reflective search~\citep{shinn2023reflexion,zhao2024expel,wang2023voyager,agrawal2025gepa}. Related memory-centric systems further study how experience can be organized and reused over longer horizons~\citep{xu2025amem,zhang2025memevolve,tang2025agent}. Closer to our setting, recent work also treats the surrounding context or harness as the optimization target: ACE evolves long-form context from execution feedback~\citep{zhang2025agentic}, Meta-Harness searches over harness code using execution traces of prior candidates~\citep{lee2026metaharness}, and AutoHarness synthesizes code harnesses that constrain agent actions in interactive environments~\citep{lou2026autoharness}.

A complementary line seeks finer-grained signals than end-of-trajectory outcomes, for example through step-level verifiers, process reward models, or intermediate rewards within a single rollout~\citep{lightman2024verify,wang2024mathshepherd,wang2025sparl,da2025agentrlvr,li2025rltr}. Our work is closest in spirit to these directions in that it also uses process-level feedback, but the signal here is temporal rather than intra-rollout: it is derived from contrasts across repeated revisits to the same unresolved question before resolution.

\section{Method}
\label{sec:method}

\emph{Milkyway} is a future prediction system that adapts before resolution. In the setting we study, the same unresolved question may be revisited at multiple checkpoints before its outcome is known, and each checkpoint may use only information available by that time. Across these revisits, \emph{Milkyway} extracts \emph{pre-resolution signals} from temporal contrasts in fixed-schema checkpoint notes and converts them into bounded updates to a typed persistent question-local state, the \emph{future prediction harness}, so that later checkpoints on the same question can benefit before the realized outcome is observed. Checkpoint notes and other runtime artifacts remain checkpoint-local, while only reusable procedural guidance is carried forward. The remainder of this section formalizes the setting (Sec.~\ref{sec:method_setting}), describes the architecture (Sec.~\ref{sec:method_overview}), and specifies the update semantics (Sec.~\ref{sec:method_feedback}).

\subsection{Future prediction setting}
\label{sec:method_setting}

To simplify notation, we consider a single future prediction question $q$. Unlike a fixed-answer task, $q$ remains unresolved at prediction time: it is issued at time $i$, and its realized outcome $y \in \mathcal{Y}$ becomes available only at a later time $r > i$. Between $i$ and $r$, the same unresolved question may be revisited at multiple time points,
\begin{equation}
i = \tau_1 < \tau_2 < \cdots < \tau_T < r,
\end{equation}
which we call checkpoints. Such repeated revisits are a property of the evaluation protocol considered here, in which each checkpoint may use only information available by that time; they are not an additional intervention introduced by Milkyway.

At checkpoint $\tau_t$, the agent may use only public information available by $\tau_t$. Under its current prediction procedure, it produces a prediction
\begin{equation}
z_t = \pi(q, \tau_t), \qquad t = 1, \dots, T.
\label{eq:method_prediction}
\end{equation}
The ordered predictions define a pre-resolution trajectory
\begin{equation}
\mathcal{T}(q) = \left[(\tau_t, z_t)\right]_{t=1}^{T},
\label{eq:method_trajectory}
\end{equation}
which records how the agent's judgment evolves while the question remains unresolved. Each checkpoint also yields a fixed-schema checkpoint note $n_t$, introduced in Sec.~\ref{sec:method_overview}, so that later checkpoints can be compared along stable dimensions.

This checkpoint structure matters because public evidence may evolve between question release and resolution, whereas outcome supervision arrives only after resolution and is coarse: the realized outcome indicates whether a prediction was correct, but not directly what the agent should have tracked, verified, or kept uncertain along the way. Once $q$ resolves, the realized outcome $y$ becomes available, and previous predictions can be evaluated with a task-specific loss $\ell(z_t, y)$.

\begin{figure*}[t]
    \centering
    \includegraphics[width=\textwidth]{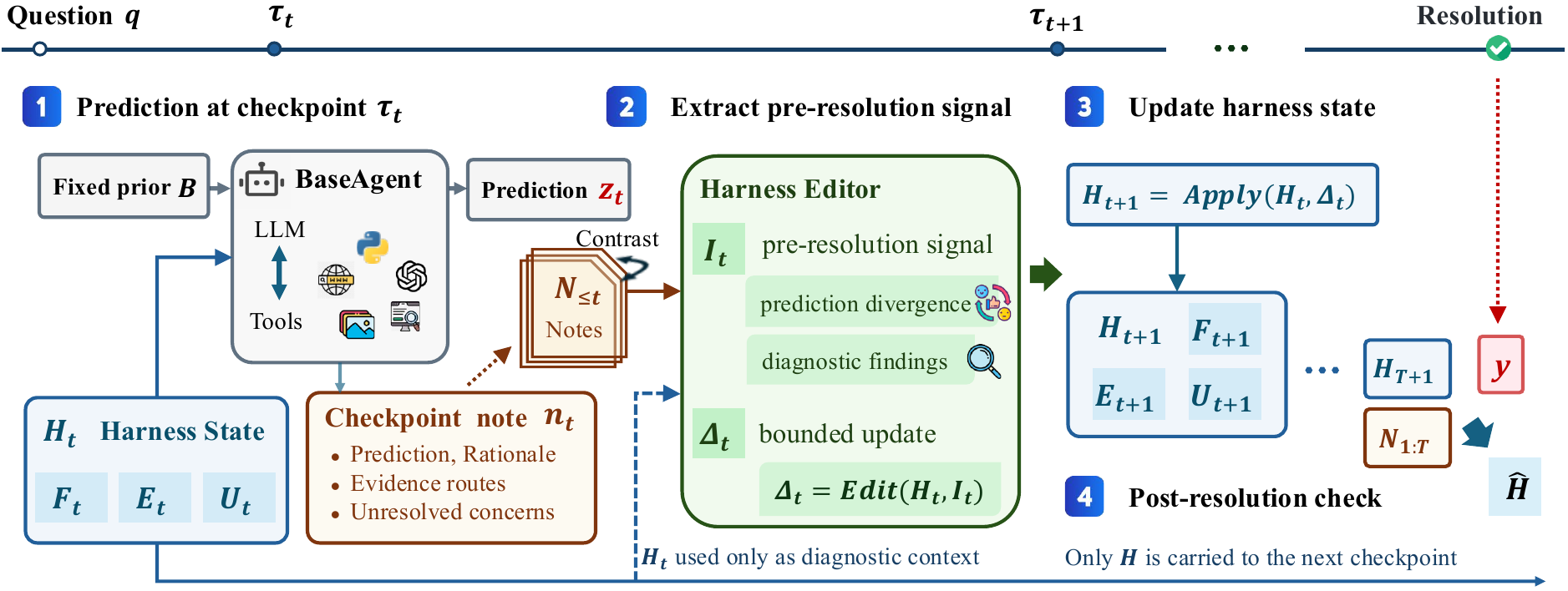}
    \caption{\textbf{Milkyway overview.} (1) At checkpoint $\tau_t$, the BaseAgent predicts under fixed prior $B$ and harness $H_t$, producing prediction $z_t$ and a fixed-schema checkpoint note $n_t$. (2) Across revisits to the same unresolved question, the ordered note history $N_{\le t}$ is compared to derive a pre-resolution signal $I_t$, which the Harness Editor maps to a bounded harness update $\Delta_t$. (3) Applying $\Delta_t$ yields $H_{t+1}$, the only state carried to the next checkpoint. (4) After resolution, the realized outcome supports a post-resolution check, yielding the checked harness $\hat H$.}
    \label{fig:method_overview}
\end{figure*}

\subsection{Milkyway: a persistent harness for future prediction}
\label{sec:method_overview}

\emph{Milkyway} is organized around a persistent question-local future prediction harness $H_t$, the only persistent state carried across checkpoints of the same unresolved question. At each checkpoint, the BaseAgent predicts under a fixed prior and harness; across revisits, a separate editing stage reads checkpoint-local notes and writes only bounded reusable guidance from pre-resolution signals into the harness.\footnote{Here, \emph{bounded} means that the editor cannot append arbitrary memory to $H_t$: it may write only reusable procedural guidance to axes $F/E/U$ via a small validated patch of \emph{add}, \emph{revise}, \emph{deprecate}, or null operations. Exact capacity and validation rules are deferred to App.~\ref{sec:app_signal_runtime}.} Figure~\ref{fig:method_overview} summarizes the architecture. The rest of this subsection defines the harness interface, the state boundary, and the admissible content of harness updates.

\paragraph{Harness interface.}
At checkpoint $\tau_t$, prediction is conditioned on two inputs: a fixed prior $B$ and a typed~harness~$H_t$,
\begin{equation}
H_t = (F_t, E_t, U_t),
\end{equation}
where $F_t$, $E_t$, and $U_t$ respectively store guidance for factor tracking, evidence handling, and uncertainty management. These axes determine what should be monitored, how evidence should be sought or verified, and when confidence should remain bounded while the question remains unresolved. In the current implementation, each active entry takes the form of a short $(\text{axis}, \text{when}, \text{guidance})$ record. The harness is thus a structured interface for question-local procedural adaptation, rather than a free-form memory buffer or a generic skill store. It is currently instantiated as structured textual entries presented to the predictor at the next checkpoint, but this textual presentation is only the current implementation of the harness, not its definition.

\paragraph{State boundary and checkpoint notes.}
Here and below, \emph{persistent state} refers to question-local information that is carried forward and made available at later checkpoints of the same unresolved question. In \emph{Milkyway}, this persistent state is exactly the harness $H_t$. At checkpoint $\tau_t$, the BaseAgent produces a prediction $z_t$ and a fixed-schema checkpoint note $n_t$. The ordered note history up to checkpoint $\tau_t$ is denoted by
\begin{equation}
N_{\le t} = (n_1, \ldots, n_t).
\end{equation}
Each checkpoint note summarizes the run in a comparable form, including the current prediction, its rationale, evidence or verification routes, and unresolved concerns. Across revisits, the Harness Editor may read $N_{\le t}$ and other checkpoint-local artifacts to diagnose temporal contrasts, but these artifacts are not themselves carried forward as persistent state. Only the resulting harness is made available at the next checkpoint.

\paragraph{Stored content.}
The harness stores only reusable procedural guidance. Such guidance may specify tracked factors, source or verification priorities, interpretations of resolution criteria, recurring failure modes, or when confidence should remain conservative. It excludes predicted answers, raw evidence, and checkpoint-specific facts. Sec.~\ref{sec:method_feedback} formalizes how note history yields pre-resolution signals and bounded updates to $H_t$.

\subsection{Pre-resolution update and post-resolution check}
\label{sec:method_feedback}

This subsection specifies the update semantics induced by the architecture above. Given the ordered note history $N_{\le t}$ and harness $H_t$, \emph{Milkyway} derives a pre-resolution signal, maps it to a bounded harness update, and after resolution checks the updated harness against the outcome.

\paragraph{Pre-resolution signal.}
For $t \ge 2$, \emph{Milkyway} compares the current checkpoint note $n_t$ with earlier notes in $N_{\le t}$ along the fixed fields of the note schema. This comparison diagnoses what the unresolved trajectory has already made visible before resolution: which factors earlier checkpoints failed to track, which evidence should have been sought or verified differently, and where confidence should have remained bounded. We define the resulting pre-resolution signal as
\begin{equation}
I_t = \Psi(H_t, N_{\le t}), \qquad t = 2, \dots, T.
\label{eq:method_signal}
\end{equation}
Here $H_t$ serves only as diagnostic context: it helps determine whether an exposed lesson is new, already covered, or inconsistent with an existing entry. In the current implementation, $I_t$ has two components: \emph{prediction divergence}, which localizes where predictions change across checkpoints and where their supporting rationales materially differ, and \emph{diagnostic findings}, which state the procedural lesson implied by those contrasts. The signal is a procedural diagnosis, not an answer to the question, and axis assignment to $F/E/U$ occurs only when a reusable lesson is written into a harness update. App.~\ref{sec:app_signal_runtime} and App.~\ref{sec:app_prompt_editor} give the concrete runtime schema, and App.~\ref{sec:app_futureworld_cases} shows a resolved example of this signal on a real trajectory.

\paragraph{Harness update from the signal.}
The Harness Editor maps the pre-resolution signal to a bounded harness update,
\begin{equation}
\Delta_t = \mathrm{Edit}(H_t, I_t),
\label{eq:method_edit}
\end{equation}
and the next harness state is obtained by
\begin{equation}
H_{t+1} = \mathrm{Apply}(H_t, \Delta_t).
\label{eq:method_apply}
\end{equation}
Operationally, boundedness constrains both \emph{what} may be written and \emph{how often} the state may change: the runtime accepts at most one validated patch per checkpoint, rather than unconstrained iterative free-form rewriting. The update consists of a small number of add, revise, or deprecate operations over the writable harness axes. A null update is admissible when the diagnosed lesson is weak, already covered, or too checkpoint-specific to support reuse. Only $H_{t+1}$ is carried forward to checkpoint $\tau_{t+1}$; $N_{\le t}$, $I_t$, and $\Delta_t$ are transient. App.~\ref{sec:app_runtime} and Table~\ref{tab:app_impl_spec} specify the deterministic application rule, while App.~\ref{sec:app_futureworld_cases} illustrates the resulting writeback on a resolved example.

\paragraph{Post-resolution check.}
Because pre-resolution updates are written before the realized outcome is known, they are provisional. Once the question resolves, the realized outcome provides a complementary check of the updated harness,
\begin{equation}
\hat{H} = \mathrm{Check}(H_{T+1}, N_{1:T}, y),
\label{eq:method_check}
\end{equation}
which may retain, refine, or deprecate previously written guidance. This closes the feedback loop for a resolved question by auditing the provisional procedural state accumulated during repeated revisits. We keep Eq.~\ref{eq:method_check} explicit because it completes the lifecycle of a resolved question, but the quantitative evidence in this paper is concentrated on the within-question pre-resolution pathway. App.~\ref{sec:app_postresolution} clarifies the current interface and the scope that is \emph{not} separately quantified here, including later cross-question reuse.

\section{Experiments}
\label{sec:experiments}

\providecommand{\pending}{\textemdash}

We evaluate \emph{Milkyway} on two live future-prediction benchmarks and then isolate the within-question pre-resolution effect under matched rolling settings. Detailed scoring rules, schedules, controls, audits, and artifacts are deferred to App.~\ref{sec:app_futureworld_eval}.

\subsection{Experimental setup}
\label{sec:exp_setup}

\paragraph{Benchmarks.}
\textsc{FutureX}~\citep{zeng2025futurex} is a weekly live benchmark with submissions on Wednesdays and four difficulty levels (L1--L4): single-choice questions with small option sets, multi-answer wide-search questions, low-volatility open-ended questions, and high-volatility open-ended questions. We use the official March~2026 Week~3 scorer and report per-level scores together with the published weighted overall $\mathrm{Ovr.}=10\%\!\cdot\!\mathrm{L1}+20\%\!\cdot\!\mathrm{L2}+30\%\!\cdot\!\mathrm{L3}+40\%\!\cdot\!\mathrm{L4}$. \textsc{FutureWorld}~\citep{FutureWorld2026han} releases 50 questions per day across binary, simple multiple-choice, difficult multiple-choice, and numerical formats. Our primary \textsc{FutureWorld} score is the simple mean over resolved questions, with type-wise means reported only for diagnosis. We report scored-set sizes in Table~\ref{tab:prediction_results}; full scoring rules and resolved-set construction are in App.~\ref{sec:app_futureworld_scoring}.

\paragraph{Forward-only protocol.}
At checkpoint $\tau_t$, a run may use only public information available by $\tau_t$; realized outcomes are used only for scoring after resolution. Main benchmark results are computed on the resolved subset of each released question pool. Any reusable state that depends on feedback is built only from resolved data preceding the scored slice and is frozen before scoring. For self-evolving baselines, we provide benchmark-matched warm-up feedback from earlier resolved questions (the immediately preceding resolved week on \textsc{FutureX}; the five resolved days preceding the scored slice on \textsc{FutureWorld}) and then freeze the evolved state during scoring. In the mechanism study, we revisit the same launched questions at matched pre-resolution horizons while fixing the backbone, scaffold, tool access, context budget, schedule, and per-checkpoint tool budget, so that only question-local persistent state differs. No method sees outcomes from the scored questions before scoring.

\paragraph{Backbones, scaffold, and baselines.}
The primary backbone is GPT-5.4; Qwen3.5-397B-A17B (Qwen3.5) and GLM-5.1 are used for cross-backbone reproducibility. All Milkyway conditions share the same worker scaffold, tool interface, effective context budget, and per-checkpoint cap of $50$ tool calls per question; in the mechanism study, conditions differ only in question-local persistent state. To reduce capacity-driven confounds, Milkyway and the baselines that we rerun use a shared effective context cap of $200$K tokens. We compare against two groups. \emph{Single-run agents}---GPT-5.4 + Search, smolagents~\citep{hf2024smolagents}, MiroFlow~\citep{su2026miroflow}, Flash-Searcher~\citep{qin2025flash}, and OpenClaw~\citep{openclaw2026}---provide strong search-and-scaffold baselines with no question-local persistence across checkpoints. \emph{Self-evolving frameworks} use reusable external state but differ in what they evolve: ACE~\citep{zhang2025agentic} evolves long-form execution context or playbooks from prior feedback; Agent KB~\citep{tang2025agent} retrieves reusable cross-task experience from a structured knowledge base; and MemEvolve~\citep{zhang2025memevolve} jointly evolves memory content and memory organization. Each is paired with the strongest compatible base agent in its public setup, yielding the ACE+smolagents, Agent KB+smolagents, and MemEvolve+Flash-Searcher rows in Table~\ref{tab:prediction_results}. Because these methods are designed to improve from prior feedback or accumulated experience, we warm-start them only on resolved questions preceding the scored slice and freeze the resulting evolved state during scoring. None implements the typed pre-resolution $F/E/U$ harness studied here. We do not compare to outcome-supervised training methods~\citep{turtel2025outcome,turtel2026futureaslabel}, which update model weights from resolved outcomes rather than reusable external state at evaluation time.

\subsection{Main benchmark results}
\label{sec:prediction_results}

\begin{table*}[!t]
\caption{\textbf{Main benchmark results on \textsc{FutureX} and \textsc{FutureWorld}.} \textsc{FutureX} uses the official March~2026 Week~3 scorer with the published weighted overall. \textsc{FutureWorld} reports the resolved five-day subset from May~2--6, 2026 ($216$ scored questions: $44$ binary, $49$ simple multiple-choice, $22$ difficult multiple-choice, and $101$ numerical); $S_{\mathrm{overall}}$ is the simple mean over all resolved questions, while type-level means are shown only for diagnosis. Milkyway is evaluated in its deployed question-level loop: each scored question is revisited three times under the forward-only protocol, and the last checkpoint is reported. Self-evolving baselines are warm-started only on resolved questions preceding the scored slice and are frozen during scoring; Sec.~\ref{sec:exp_setup} gives the benchmark-specific warm-up details. $^\dagger$ marks officially released \textsc{FutureX} numbers.}
\label{tab:prediction_results}
\centering
\small
\renewcommand{\arraystretch}{1.15}
\setlength{\tabcolsep}{4.2pt}
\resizebox{\textwidth}{!}{%
\begin{tabular}{@{}llccccc@{\hspace{8pt}}ccccc@{}}
\toprule
\multirow{2}{*}[-0.8ex]{\normalsize\textbf{Method}} & \multirow{2}{*}[-0.8ex]{\normalsize\textbf{Backbone}} &
\multicolumn{5}{c}{\raisebox{-0.2ex}[2.4ex][0.8ex]{\textbf{\textsc{FutureX}$\uparrow$}}} &
\multicolumn{5}{c}{\raisebox{-0.2ex}[2.4ex][0.8ex]{\textbf{\textsc{FutureWorld}$\uparrow$}}} \\
\cmidrule(lr){3-7}\cmidrule(l){8-12}
& & L1 & L2 & L3 & L4 & Ovr. & $S_{\mathrm{bin}}$ & $S_{\mathrm{smc}}$ & $S_{\mathrm{dmc}}$ & $S_{\mathrm{num}}$ & $S_{\mathrm{overall}}$ \\[0.2ex]
\midrule
GPT-5.4 + Search & GPT-5.4 & 62.14 & 59.80 & 44.24 & 31.57 & 44.07 & 72.43 & 52.86 & 41.74 & 45.71 & 52.37 \\
\midrule
\rowcolor{black!10}\multicolumn{12}{@{}l}{\textit{Single-run agents}} \\
smolagents & GPT-5.4 & 63.20 & 61.35 & 46.10 & 34.25 & 46.12 & 75.28 & 55.41 & 43.27 & 49.18 & 55.31 \\
MiroFlow$^\dagger$ & GPT-5.4 & 64.29 & 72.82 & 59.45 & \textbf{46.80} & 57.55 & \textbf{90.76} & 63.42 & 56.37 & 59.14 & 66.27 \\
Flash-Searcher & GPT-5.4 & 65.10 & 67.35 & 52.20 & 39.80 & 51.56 & 78.64 & \textbf{70.18} & 48.73 & 52.88 & 61.63 \\
OpenClaw & GPT-5.4 & 64.00 & 60.80 & 45.80 & 34.00 & 45.90 & 67.62 & 58.47 & 40.28 & 46.62 & 52.94 \\
\midrule
\rowcolor{black!10}\multicolumn{12}{@{}l}{\textit{Self-evolving agents}} \\
ACE + smolagents & GPT-5.4 & 65.95 & 68.20 & 53.05 & 40.65 & 52.41 & 80.31 & 59.12 & 49.37 & 53.45 & 59.79 \\
Agent KB + smolagents & GPT-5.4 & 66.35 & 68.85 & 53.60 & 41.25 & 52.99 & 81.24 & 59.63 & 50.21 & 53.91 & 60.40 \\
MemEvolve + Flash-Searcher & GPT-5.4 & 67.05 & 69.60 & 54.45 & 42.10 & 53.80 & 82.17 & 60.28 & 50.64 & 54.04 & 60.84 \\
\midrule
\rowcolor{black!10}\multicolumn{12}{@{}l}{\textit{Ours}} \\
\textbf{Milkyway}$^\dagger$ & GPT-5.4 & \textbf{71.43} & \textbf{82.26} & \textbf{63.05} & 45.85 & \textbf{60.85} & 88.76 & 69.54 & \textbf{62.31} & \textbf{61.69} & \textbf{69.05} \\
\textbf{Milkyway} & GLM-5.1 & 68.20 & 70.45 & 55.10 & 43.20 & 54.72 & 83.74 & 60.36 & 52.19 & 55.83 & 62.17 \\
\textbf{Milkyway} & Qwen3.5 & 67.70 & 70.12 & 54.62 & 43.28 & 54.49 & 83.21 & 60.05 & 51.86 & 55.47 & 61.79 \\
\bottomrule
\end{tabular}
}
\end{table*}

Table~\ref{tab:prediction_results} is an end-to-end comparison under each method's evaluation-time protocol rather than a component ablation. On the scored slice, Milkyway performs within-question pre-resolution writeback across revisits and reports the final pre-resolution checkpoint. Post-resolution checking remains part of the lifecycle, but in the current implementation it audits guidance only at the question level; this table therefore does not isolate whether checked guidance later helps other questions. For self-evolving baselines, we provide only warm-up from resolved data preceding the scored slice and freeze the resulting state during scoring. Thus, no method sees outcomes from the scored questions before scoring, and no self-evolving baseline continues evolving on the evaluation slice.

Under this protocol, Milkyway with GPT-5.4 attains the best \textsc{FutureX} overall score and leads on L1--L3. On \textsc{FutureWorld}, it achieves the strongest overall score under the simple mean over resolved questions. The win is not driven by uniformly topping every type: MiroFlow remains strongest on binary questions and Flash-Searcher on simple multiple-choice, but Milkyway's larger margins on difficult multiple-choice and numerical questions more than offset those gaps. Relative to the strongest self-evolving baseline, Milkyway improves the overall score by $+7.05$ on \textsc{FutureX} and $+8.21$ on \textsc{FutureWorld}; even against the best single-run baseline, the margins remain $+3.30$ and $+2.78$, respectively. The GLM-5.1 and Qwen3.5 rows further show that the same deployment recipe is not specific to a single backbone.

\subsection{Mechanism study}
\label{sec:exp_mechanism}

\begin{table}[t]
\caption{\textbf{Matched-state mechanism readout on rolling \textsc{FutureWorld} cells (T-4d $\to$ T-1d).} $\Delta_C = S_C(\text{T-1d}) - S_C(\text{T-4d})$ measures within-cell improvement from the earliest to the final pre-resolution checkpoint. Within each row, backbone, scaffold, tool access, launched questions, daily schedule, resolved subset, effective context budget, and the $50$-tool-call cap are fixed; only question-local persistent state differs. $n$ is the matched resolved-set size. ``---'' denotes matched-budget \textsc{GH} runs not yet completed on the open-source backbones.}
\label{tab:mechanism_readout}
\centering
\small
\renewcommand{\arraystretch}{1.10}
\setlength{\tabcolsep}{5.0pt}
\begin{tabular}{lccccccc}
\toprule
\textbf{Backbone} & \textbf{Date} & $n$
& $\boldsymbol{\Delta_{\textsc{NH}}}$
& $\boldsymbol{\Delta_{\textsc{GH}}}$
& $\boldsymbol{\Delta_{\textsc{FH}}}$
& $\boldsymbol{\Delta_{\textsc{FH}}{-}\Delta_{\textsc{NH}}}$
& $\boldsymbol{\Delta_{\textsc{FH}}{-}\Delta_{\textsc{GH}}}$ \\
\midrule
GPT-5.4 & 2026-05-05 & 35 & $+0.9$ & $+6.0$ & $+14.0$ & $\mathbf{+13.1}$ & $\mathbf{+8.0}$ \\
GPT-5.4 & 2026-05-06 & 28 & $+7.8$ & $+7.8$ & $+16.9$ & $\mathbf{+9.1}$  & $\mathbf{+9.1}$ \\
Qwen3.5 & 2026-05-06 & 28 & $+3.7$ & ---     & $+13.0$ & $\mathbf{+9.3}$  & --- \\
GLM-5.1 & 2026-05-06 & 28 & $+4.5$ & ---     & $+15.5$ & $\mathbf{+11.0}$ & --- \\
\bottomrule
\end{tabular}
\end{table}

\begin{figure*}[!tbp]
\centering
\includegraphics[width=\textwidth]{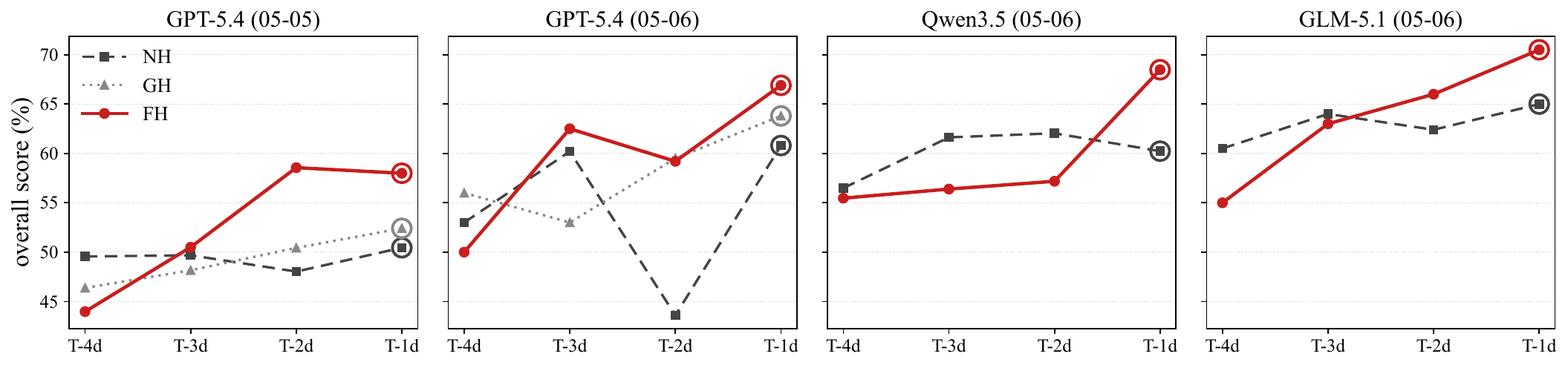}
\caption{\textbf{Pre-resolution score trajectories on four matched cells.} The panels show GPT-5.4 / 05-05, GPT-5.4 / 05-06, Qwen3.5 / 05-06, and GLM-5.1 / 05-06 (left to right). \textsc{NH} removes question-local persistence; \textsc{GH} keeps persistence with a free-form memory blob under matched byte and write-call budgets; \textsc{FH} uses the typed $F/E/U$ harness. Because non-empty writeback cannot be used before T-2d, the earlier checkpoints are pre-writeback controls. \textsc{FH} separates only after writeback becomes available and achieves the largest T-4d$\rightarrow$T-1d gain in every cell.}
\label{fig:mechanism_trajectories}
\end{figure*}

Unlike the self-evolving baselines in Table~\ref{tab:prediction_results}, which are frozen during the scored slice after pre-evaluation warm-up, the conditions here are constructed to test whether within-question state updates during the scored pre-resolution trajectory improve forecasting. We build a matched-state comparison on rolling \textsc{FutureWorld} questions from T-4d to T-1d. Within each cell, we fix the backbone, scaffold, tool access, launched questions, daily schedule, resolved subset, effective context budget, and the per-checkpoint $50$-tool-call cap; only the form of question-local persistent state changes. \textsc{NH} removes persistence and measures repeated forecasting alone. \textsc{GH} keeps persistence but replaces the typed harness with a free-form memory blob under matched byte and write-call budgets. \textsc{FH} uses the typed $F/E/U$ harness with bounded edits. Because the editor needs at least two earlier checkpoint notes, no condition can consume non-empty writeback before T-2d; the earlier checkpoints therefore serve as built-in pre-writeback controls.

Table~\ref{tab:mechanism_readout} reports the endpoint improvement $\Delta_C = S_C(\text{T-1d}) - S_C(\text{T-4d})$, and Fig.~\ref{fig:mechanism_trajectories} shows the full trajectories. The predicted signature is delayed separation rather than an immediate offset: if typed pre-resolution writeback is the driver, \textsc{FH} should pull away only after writeback can affect the next run. This is the pattern observed in all four matched cells. \textsc{FH} yields the largest T-4d$\rightarrow$T-1d improvement in every cell, exceeding \textsc{NH} by $+13.1$, $+9.1$, $+9.3$, and $+11.0$ points. Where matched-budget \textsc{GH} is available, \textsc{FH} also exceeds generic persistence by $+8.0$ and $+9.1$ points. Repeated forecasting under matched compute therefore explains only part of the rise, and free-form persistence does not recover the effect of typed writeback. Supporting controls in App.~\ref{sec:app_reporting_surfaces} and App.~\ref{sec:app_futureworld_artifacts} further bound same-day reruns without calendar-time evidence, extra tool or token budget, and generic memory accumulation.

\begin{table}[!t]
\caption{\textbf{Case study: typed writeback changes the evidence route, not the answer.} Gold $=4.48$. A Hebei province-level \emph{inbound} migration question. Typed writes correct the evidence route, while \textsc{NH}/\textsc{GH} stay on the wrong metric under the same scaffold and tool budget. Verbatim writes and the contrasting \textsc{GH} blob appear in App.~Fig.~\ref{fig:case_study}.}
\label{tab:case_summary}
\centering
\footnotesize
\renewcommand{\arraystretch}{1.12}
\setlength{\tabcolsep}{3.6pt}
\begin{tabularx}{\columnwidth}{@{}c c >{\raggedright\arraybackslash}X rrr@{}}
\toprule
\multirow{2}{*}{\textbf{Ckpt}} &
\multirow{2}{*}{\textbf{Axis}} &
\multirow{2}{*}{\textbf{Typed writeback / active state}} &
\multicolumn{3}{c@{}}{\textbf{Predictions}} \\
\cmidrule(l){4-6}
& & & \textbf{\textsc{FH}} & \textbf{\textsc{NH}} & \textbf{\textsc{GH}} \\
\midrule
T-4d & --- &
Empty harness; the run anchors on a wrong-surface ranking statistic rather than the target metric.
& 35.20 & 0.98 & 27.10 \\

T-3d & $U$ &
No direct same-metric point is observed; cap confidence and wait for a direct anchor.
& 5.46 & 33.60 & 28.10 \\

T-2d & $E$ &
Rankings and legacy endpoints are wrong surfaces; route first to the live inbound-destination province view.
& 5.23 & 27.50 & 27.10 \\

T-1d & $U{+}E$ &
Reuse the typed state; the corrected route is retained and no new write is needed.
& \textbf{5.24} & 27.18 & 27.30 \\
\bottomrule
\end{tabularx}
\end{table}

\subsection{Case study}
\label{sec:exp_case}

Table~\ref{tab:case_summary} makes the writeback pathway concrete on a resolved Hebei province-level \emph{inbound} migration question (gold $4.48$). The initial failure is metric identification: the run reads a wrong-surface statistic around $27$ rather than the target metric. The ordered notes then trigger two reusable writes: a $U$ entry that keeps confidence bounded until a direct same-metric anchor is found, and an $E$ entry that redirects later runs to the live Baidu Migration UI under the province-level inbound-destination view. Once these typed entries become active, \textsc{FH} stays near the gold ($5.46 \rightarrow 5.23 \rightarrow 5.24$), whereas the matched T-1d \textsc{NH}/\textsc{GH} controls remain on the wrong surface at $27.18/27.30$. The case therefore illustrates the mechanism claimed in Sec.~\ref{sec:exp_mechanism}: the harness stores a reusable evidence-handling procedure, not the answer itself. Full artifacts and verbatim writes are in App.~\ref{sec:app_futureworld_cases}.

\section{Discussion and conclusion}
\label{sec:discussion_conclusion}

Milkyway is intentionally limited to typed, question-local harness updates within a single unresolved question. It does not modify tools, maintain reusable datasets, reorganize broader workflow state, or alter the agent scaffold. Thus, the results support within-question pre-resolution adaptation, not fully general long-horizon self-evolution. Post-resolution checking remains part of the lifecycle, but here it audits guidance only at the question level; cross-question transfer is not separately quantified.

Within this scope, the results support the central claim: unresolved trajectories can provide useful pre-resolution signals. By extracting procedural lessons from temporal contrasts across revisits and writing them back into the future prediction harness, \emph{Milkyway} improves later forecasts before realized outcomes are available. Future work should study richer adaptation surfaces and longer-horizon interactions among pre-resolution writeback, post-resolution checking, and cross-question transfer.

\clearpage
\bibliographystyle{plainnat}
\bibliography{refs}

\clearpage
\appendix
\section*{Appendix}
\addcontentsline{toc}{section}{Appendix}
\appendix

\section{Method details}
\label{sec:app_impl}

This appendix records the runtime realization of the objects introduced in Sec.~\ref{sec:method_overview} and Sec.~\ref{sec:method_feedback}. The main text defines the architecture and update semantics; here we specify only the runtime objects, the editor return, the deterministic application rule, and the within-question loop used in the experiments.

\subsection{Runtime objects}
\label{sec:app_state}

At checkpoint $\tau_t$, the BaseAgent runs under the fixed prior $B$ and the current harness $H_t$. The run terminates through a structured finalization action that returns a finalized payload $p_t$, from which the prediction $z_t$ is read out. The full runtime trace $r_t$ is retained only as a checkpoint-local artifact for diagnosis. After the run is complete, the checkpoint note is constructed post hoc as
\begin{equation}
n_t = \Gamma(r_t, p_t).
\label{eq:app_note_construction}
\end{equation}
This construction does not affect the prediction at $\tau_t$; it only converts a completed run into a compact artifact for later temporal comparison.

The state boundary from Sec.~\ref{sec:method_overview} is enforced directly at runtime. The harness $H_t$ is the only persistent question-local state carried across checkpoints. The trace $r_t$, payload $p_t$, and checkpoint note $n_t$ are checkpoint-local artifacts: they may be inspected when constructing or validating an update, but they are not themselves carried forward as persistent state.

\paragraph{Checkpoint-note schema and content constraint.}
In the current implementation, every checkpoint note follows the same fixed schema. It records the current prediction, its supporting and countervailing rationale, evidence or verification routes, unresolved concerns, and an optional non-load-bearing reflection field. The core signal semantics use the fixed fields rather than the optional reflection. This schema makes note history comparable across checkpoints: later diagnosis can contrast how the prediction changed, how evidence handling changed, and which unresolved concerns persisted.

The content restriction from Sec.~\ref{sec:method_overview} is enforced at validation time. A candidate harness entry is admissible only if it expresses reusable procedural guidance for later checkpoints on the same unresolved question. Examples include factors that should be tracked, source or verification priorities, interpretations of resolution criteria, recurring failure modes with recommended alternatives, and fallback or verification routes that remain useful across later checkpoints. Predicted answers, raw evidence dumps, and checkpoint-specific facts are rejected.

\paragraph{Minimal concrete example.}
A checkpoint note can be read as a compact record of the form: current prediction, supporting and countervailing rationale, evidence route, and unresolved concern. A harness entry then distills only the reusable lesson exposed by contrasts across such notes, for example: \emph{axis $E$; when related pages expose a topic-relevant but metric-mismatched surface; guidance route first to the official surface that matches the benchmark metric and treat ranking pages only as background}. This keeps the persistent state procedural rather than factual. App.~\ref{sec:app_futureworld_cases} shows the full resolved artifact behind this abstraction.

\subsection{Editor return and bounded update}
\label{sec:app_signal_runtime}

For $t \ge 2$, the runtime packages the current harness $H_t$ and the ordered note history $N_{\le t}=(n_1,\ldots,n_t)$ into a single editor input. The editor makes one structured call that returns both the pre-resolution signal
\[
I_t = \Psi(H_t, N_{\le t})
\]
and the bounded harness update
\[
\Delta_t = \mathrm{Edit}(H_t, I_t).
\]
They are returned together for runtime efficiency, but they remain semantically distinct: $I_t$ is a diagnosis over note history, whereas $\Delta_t$ is the proposed update to persistent harness state. At runtime, Eq.~\ref{eq:method_signal} appears as the \texttt{signal} field of a structured editor output, and Eq.~\ref{eq:method_edit} appears as the corresponding \texttt{update} field; the prompt excerpt and output schema are shown in App.~\ref{sec:app_prompt_editor}.

In the current implementation, $I_t$ has two components. \emph{Prediction divergence} localizes where predictions change across checkpoints and where their supporting rationales materially differ. \emph{Diagnostic findings} interpret those contrasts and state what should be tracked, verified, or kept uncertain going forward. The signal itself is not partitioned into $F/E/U$; axis assignment occurs only when a reusable procedural lesson is written into the harness update.

The update is bounded. Its admissible operations are \texttt{add}, \texttt{revise}, and \texttt{deprecate}, each attached to one writable harness axis in $\{F,E,U\}$. The editor may also return a null update when the diagnosed lesson is weak, already covered, or too checkpoint-specific to support reuse. This prevents every cross-checkpoint contrast from being forced into the harness.

\subsection{Application and presentation}
\label{sec:app_runtime}

Application of the bounded harness update is deterministic:
\begin{equation}
H_{t+1} = \mathrm{Apply}(H_t, \Delta_t).
\label{eq:app_update}
\end{equation}
The applier validates entry references, filters updates that are clearly redundant with active entries, applies operations in a fixed order (\texttt{deprecate} before \texttt{revise} before \texttt{add}), and degrades invalid or null proposals to no-ops. Thus, the model-side decision in the update stage ends with the proposed $\Delta_t$; the state transition itself is handled by a deterministic procedure.

Table~\ref{tab:app_impl_spec} gives the implementation-level constraints that make the bounded-update claim in Sec.~\ref{sec:method_feedback} operationally precise. Together, Eq.~\ref{eq:method_edit} proposes a typed patch and Eq.~\ref{eq:method_apply} deterministically validates and applies it.

After application, the resulting harness $H_{t+1}$ is presented to the predictor at checkpoint $\tau_{t+1}$ alongside the fixed prior $B$. In the current implementation, this presentation takes the form of a structured guidance block. This block is only the current textual realization of the harness, not the definition of the harness itself. It is used here to isolate the writeback pathway, so that differences across conditions are attributable to harness updates rather than to whether an optional artifact happened to be retrieved or reread.

\begin{table}[tbp]
\caption{\textbf{Implementation constraints for bounded harness updates.} These constraints make the bounded-update claim in Sec.~\ref{sec:method_feedback} operational: the editor proposes a small typed patch, and the runtime either applies a validated version of it or degrades it to a no-op.}
\label{tab:app_impl_spec}
\centering
\small
\renewcommand{\arraystretch}{1.10}
\setlength{\tabcolsep}{5pt}
\begin{tabular}{@{}p{0.28\textwidth}p{0.16\textwidth}p{0.42\textwidth}@{}}
\toprule
\textbf{Parameter} & \textbf{Value} & \textbf{Role at validation/application} \\
\midrule
Admissible ops & \texttt{add}, \texttt{revise}, \texttt{deprecate}, null & Operations outside this set are rejected. Null is allowed when the diagnosed lesson is weak or non-reusable. \\
Patches per checkpoint & at most one & Prevents iterative free-form rewriting within a single checkpoint. \\
New adds per patch & at most 2 & Keeps each writeback local to a small number of reusable lessons. \\
Per-axis active-entry cap & 5 entries & Additional adds beyond the cap are dropped rather than expanding the state unboundedly. \\
Duplicate or invalid edits & filtered at validation & Near-duplicate adds and bad revise/deprecate references degrade to no-ops. \\
Application order & \texttt{deprecate} $\to$ \texttt{revise} $\to$ \texttt{add} & Fixed across checkpoints so that the state transition is deterministic once $\Delta_t$ is proposed. \\
\bottomrule
\end{tabular}
\end{table}

\subsection{Runtime loop}
\label{sec:app_loop}

Algorithm~\ref{alg:app_preresolution} summarizes the within-question pre-resolution loop used in the main experiments. The mechanism ablation initializes $H_1$ as empty. In the benchmark comparison, $H_1$ is simply the initial harness available at the first scored revisit under the deployed protocol; the paper does not separately quantify any cross-question benefit from post-resolution checking.

\begin{algorithm}[t]
\caption{Within-question pre-resolution adaptation in \emph{Milkyway}.}
\label{alg:app_preresolution}
\small
\begin{algorithmic}[1]
\Require unresolved question $q$, checkpoints $\tau_1 < \cdots < \tau_T$, fixed prior $B$, initial harness $H_1$
\For{$t = 1,\dots,T$}
    \State run the BaseAgent under $(B, H_t)$ to obtain $(r_t, p_t)$ and read out $z_t$ from $p_t$
    \State construct checkpoint note $n_t = \Gamma(r_t, p_t)$
    \If{$t \ge 2$}
        \State package $(H_t, N_{\le t})$ and call the editor once to obtain $(I_t, \Delta_t)$
        \State $H_{t+1} \gets \mathrm{Apply}(H_t, \Delta_t)$
    \Else
        \State $H_{t+1} \gets H_t$
    \EndIf
    \If{$t < T$}
        \State present $H_{t+1}$ to the predictor at checkpoint $\tau_{t+1}$
    \EndIf
\EndFor
\end{algorithmic}
\end{algorithm}

\subsection{Post-resolution check}
\label{sec:app_postresolution}

After resolution, the realized outcome $y$ supports the complementary check
\[
\hat{H} = \mathrm{Check}(H_{T+1}, N_{1:T}, y)
\]
from Eq.~\ref{eq:method_check}. We write this interface to denote the lifecycle endpoint of a resolved question: in principle it can retain, refine, or deprecate provisional guidance in light of the realized outcome. It complements the pre-resolution pathway because note history can expose provisional procedural lessons before $y$ is known, while the outcome can later determine which of those lessons should survive.

For the experiments reported here, however, the paper does not separately ablate this stage or quantify later cross-question reuse of checked guidance. The quantitative evidence should therefore be read as support for the within-question pre-resolution pathway, with Eq.~\ref{eq:method_check} included for completeness of the lifecycle rather than as a separately isolated gain claim.

\section{\textsc{FutureX} protocol details}
\label{sec:app_futurex_protocol}

\textsc{FutureX} is evaluated under its weekly submission protocol. Systems submit on Wednesday, and the submitted questions cover events whose resolution windows fall over the following week. For the March~2026 Week~3 slice used in Tab.~\ref{tab:prediction_results}, target questions are revisited on Monday, Tuesday, and Wednesday before submission; the Monday and Tuesday runs provide the pre-resolution trajectory, and the Wednesday run is the reported prediction submitted for scoring.

The same forward-only boundary applies as in the \textsc{FutureWorld} experiments: at each run, the agent may use only public information available at that time, and no outcome from the target evaluation week is used before scoring. The official \textsc{FutureX} scorer reports L1--L4 and the weighted overall score used in the main table.

\section{\textsc{FutureWorld} experiment details}
\label{sec:app_futureworld_eval}

\providecommand{\pending}{\textemdash}

This appendix provides the scorer definition, scored-set construction and cohort schedules, baseline provenance, full rolling ladders, the same-day repeated-rounds control, the per-question compute footprint, the artifact-level mechanism audit, and qualitative cases referenced by Sec.~\ref{sec:experiments}.

\subsection{\textsc{FutureWorld} scorer}
\label{sec:app_futureworld_scoring}

Only resolved \textsc{FutureWorld} questions with valid retrieved ground truth are scored; unresolved questions are excluded from the mean. The main table reports type-level scores for diagnosis and an overall score defined as the simple mean of per-question scores over the resolved set.

For binary choice, simple multiple choice, and difficult multiple choice questions, the gold answer and model prediction are mapped to binary indicator vectors over the candidate options. For a question with $m$ options, let $\mathbf{y}, \hat{\mathbf{y}} \in \{0,1\}^m$ denote the gold and predicted vectors. The instance score is option-level F1:
\begin{equation}
s_{\mathrm{choice}}(\mathbf{y}, \hat{\mathbf{y}})
=
\frac{2\,\mathbf{y}^{\top}\hat{\mathbf{y}}}
{\|\mathbf{y}\|_1 + \|\hat{\mathbf{y}}\|_1}.
\end{equation}
For binary choice questions, selecting more than one option is invalid and receives score 0.

For numeric prediction questions, let $\hat{v}$ be the predicted value, $v$ be the resolved value, and $\mathcal{V}=\{v_1,\ldots,v_8\}$ be the target-day value plus the seven most recent historical values used by the benchmark. The score is
\begin{equation}
s_{\mathrm{num}}
=
\max\!\left(
0,\;
1 -
\left(
\frac{\hat{v} - v}
{3\sigma(\mathcal{V}) + \varepsilon}
\right)^2
\right),
\end{equation}
where $\sigma(\mathcal{V})$ is the sample standard deviation and $\varepsilon$ is a small numerical-stability constant.

Let $\mathcal{T}$ denote the full resolved scored set. The overall score reported in Tab.~\ref{tab:prediction_results} is
\begin{equation}
S_{\mathrm{overall}}
=
\frac{1}{|\mathcal{T}|}\sum_{i \in \mathcal{T}} s_i.
\end{equation}
For diagnostic transparency, we also report type-level means over the resolved subsets $\mathcal{T}_{\mathrm{bin}}$, $\mathcal{T}_{\mathrm{smc}}$, $\mathcal{T}_{\mathrm{dmc}}$, and $\mathcal{T}_{\mathrm{num}}$:
\begin{equation}
S_{k} = \frac{1}{|\mathcal{T}_k|}\sum_{i \in \mathcal{T}_k} s_i,
\quad
k \in \{\mathrm{bin},\mathrm{smc},\mathrm{dmc},\mathrm{num}\}.
\end{equation}
For the resolved five-day benchmark comparison used in Tab.~\ref{tab:prediction_results}, the scored set contains 44 binary-choice, 49 simple multiple-choice, 22 difficult multiple-choice, and 101 numerical questions. Numeric tasks for which the benchmark cannot recover the eight-point history are excluded from the corresponding resolved subset and are flagged in the per-cohort artifact ledger.

\subsection{Cohorts, checkpoints, and evaluation boundary}
\label{sec:app_futureworld_protocol}

The daily benchmark releases a live question batch before outcomes are known and retrieves ground truth after the expected resolution window. Our main \textsc{FutureWorld} comparison uses the resolved five-day slice summarized in Tab.~\ref{tab:prediction_results}. Across this slice, unresolved or invalid-ground-truth items are excluded, leaving 216 scored questions. Milkyway revisits each target item three times before resolution and reports the last pre-resolution prediction, one day before resolution.

The mechanism study uses selected daily cohorts with matched pre-resolution checkpoints. At checkpoint T$-k$d, the run is executed $k$ calendar days before the target resolution date; runs on or after the resolution date are excluded from the forward-only analysis. Table~\ref{tab:app_futureworld_schedule} summarizes the evidence surfaces explicitly analyzed in the appendix. The cohort ids are internal artifact identifiers; dates and roles are shown first for readability.

\begin{table}[tbp]
\caption{Eligible \textsc{FutureWorld} cohorts and their roles. All score claims are made only on pre-resolution checkpoints with a shared resolved-question filter. Internal cohort ids are retained to align the paper with released artifacts.}
\label{tab:app_futureworld_schedule}
\centering
\scriptsize
\renewcommand{\arraystretch}{1.10}
\setlength{\tabcolsep}{4pt}

\resizebox{\textwidth}{!}{%
\begin{tabular}{@{}lccp{0.34\textwidth}p{0.25\textwidth}@{}}
\toprule
\textbf{Date / cohort} & \textbf{Resolution} & \textbf{Eligible checkpoints} & \textbf{Role in the experiment} & \textbf{Used by} \\
\midrule
May 2--4 cohorts (\texttt{c52--c54}) & May 2--4 & T$-1$d & Near-resolution single-checkpoint supplements used for supporting main-slice reporting; do not test harness writeback. & Main-table support only. \\
May 5 cohort (\texttt{c55}) & May 5 & T$-4$d--T$-1$d & Robustness ladder for GPT-5.4 \textsc{NH}/\textsc{GH}/\textsc{FH}. & Tab.~\ref{tab:app_rolling_ladders}. \\
May 6 cohort (\texttt{c56}) & May 6 & T$-4$d--T$-1$d & Headline ladder for the rolling-cohort ablation and backbone replications. & Tab.~\ref{tab:app_rolling_ladders}; this appendix. \\
May 6 same-day repeat (\texttt{c56\_e2}) & May 6 & same-day rounds & Same-calendar repeated rounds; control for repeated reflection without calendar-time evidence drift. & Tab.~\ref{tab:app_supporting_controls}. \\
\bottomrule
\end{tabular}
}
\end{table}

\subsection{Baseline provenance and run configuration}
\label{sec:app_baseline_provenance}

Table~\ref{tab:app_baseline_provenance} records the source and evaluation configuration for the rows in Tab.~\ref{tab:prediction_results}. All rows obey the same forward-only boundary: public information may be used only up to the run checkpoint, and target-cohort outcomes are used only after resolution for scoring.

\begin{table}[tbp]
\caption{\textbf{Provenance of benchmark-comparison rows.} ``Official'' denotes numbers taken from a released benchmark result; ``rerun'' denotes our forward-only run on the same evaluation slice.}
\label{tab:app_baseline_provenance}
\centering
\small
\renewcommand{\arraystretch}{1.12}
\setlength{\tabcolsep}{4.0pt}

\resizebox{\textwidth}{!}{%
\begin{tabular}{@{}lp{0.15\textwidth}p{0.20\textwidth}p{0.32\textwidth}p{0.22\textwidth}@{}}
\toprule
\textbf{Row} & \textbf{Source} & \textbf{Backbone / scaffold} & \textbf{Evaluation window} & \textbf{State configuration} \\
\midrule
GPT-5.4 + Search & Rerun & GPT-5.4 with native search & \textsc{FutureX} March~2026 Week~3; \textsc{FutureWorld} resolved five-day slice used in Tab.~\ref{tab:prediction_results} & Single checkpoint; no cross-checkpoint state. \\
\midrule
\rowcolor{black!10}\multicolumn{5}{@{}l}{\textit{Single-run agent baselines}} \\
smolagents & Rerun & GPT-5.4 with smolagents scaffold & Same as above & Single checkpoint; no cross-checkpoint state. \\
MiroFlow & Official / rerun mix & GPT-5.4 with released MiroFlow scaffold & Official \textsc{FutureX} slice where available; \textsc{FutureWorld} resolved subset rerun & Released single-run framework; no Milkyway harness. \\
Flash-Searcher & Rerun & GPT-5.4 with Flash-Searcher scaffold & Same as above & Released single-run search framework; no Milkyway harness. \\
OpenClaw & Rerun & GPT-5.4 with OpenClaw scaffold & Same as above & Released single-run agent scaffold; no Milkyway harness. \\
\midrule
\rowcolor{black!10}\multicolumn{5}{@{}l}{\textit{Self-evolving agent frameworks}} \\
ACE + smolagents & Rerun & GPT-5.4; ReAct worker implemented with smolagents & Same as above & Reusable external state following ACE-style trajectory experience; no typed $F/E/U$ harness. \\
Agent KB + smolagents & Rerun & GPT-5.4; strongest compatible base scaffold & Same as above & Reusable knowledge base following Agent KB; no typed $F/E/U$ harness. \\
MemEvolve + Flash-Searcher & Rerun & GPT-5.4; strongest compatible Flash-Searcher base & Same as above & Evolved memory store following MemEvolve; no typed $F/E/U$ harness. \\
\midrule
\rowcolor{black!10}\multicolumn{5}{@{}l}{\textit{Ours}} \\
Milkyway & Rerun / official \textsc{FutureX} release where marked & GPT-5.4, GLM-5.1, or Qwen3.5-397B-A17B & Same benchmark slices as the corresponding row & Deployed question-level loop with target-cohort pre-resolution revisits; last pre-resolution run reported. \\
\bottomrule
\end{tabular}
}
\end{table}

\subsection{Reporting surfaces}
\label{sec:app_reporting_surfaces}

This subsection collects the rolling ladders, the same-day repeated-rounds control, the per-question compute footprint, and the backbone reproducibility table referenced from the main text.

\paragraph{Rolling ladders.}
Table~\ref{tab:app_rolling_ladders} reports the raw per-checkpoint trajectories underlying Tab.~\ref{tab:mechanism_readout} and Fig.~\ref{fig:mechanism_trajectories}. To keep the appendix table compact, we omit the per-checkpoint \textsc{FH}--control gaps; the endpoint gaps are already summarized in Tab.~\ref{tab:mechanism_readout}.

\begin{table}[tbp]
\caption{\textbf{Rolling ladders for the four matched cells of Fig.~\ref{fig:mechanism_trajectories}.} Each row is scored on the same matched setting at matched daily checkpoints under the forward-only protocol with the \textsc{FutureWorld} scorer of App.~\ref{sec:app_futureworld_scoring}. T-4d is the null control because the harness is empty for all conditions by construction; the editor first writes after the second daily run, so T-2d is the first checkpoint at which \textsc{FH} consumes a non-empty harness. Values match Fig.~\ref{fig:mechanism_trajectories}.}
\label{tab:app_rolling_ladders}
\centering
\small
\renewcommand{\arraystretch}{1.12}
\setlength{\tabcolsep}{6pt}
\begin{tabular}{@{}llccccc@{}}
\toprule
\textbf{Setting} & \textbf{Cond.} & \textbf{T-4d} & \textbf{T-3d} & \textbf{T-2d} & \textbf{T-1d} & $\boldsymbol{\Delta_{\text{T-1d}-\text{T-4d}}}$ \\
\midrule
\multicolumn{7}{@{}l}{\textit{GPT-5.4 / 05-05 ($n=35$)}} \\
& \textsc{NH} & 49.6 & 49.7 & 48.0 & 50.4 & $+0.9$ \\
& \textsc{GH} & 46.4 & 48.2 & 50.4 & 52.4 & $+6.0$ \\
& \textsc{FH} & 44.0 & 50.5 & 58.6 & 58.0 & $+14.0$ \\
\addlinespace[3pt]
\multicolumn{7}{@{}l}{\textit{GPT-5.4 / 05-06 ($n=28$)}} \\
& \textsc{NH} & 53.0 & 60.2 & 43.6 & 60.8 & $+7.8$ \\
& \textsc{GH} & 56.0 & 53.0 & 59.5 & 63.8 & $+7.8$ \\
& \textsc{FH} & 50.0 & 62.5 & 59.2 & 66.9 & $+16.9$ \\
\addlinespace[3pt]
\multicolumn{7}{@{}l}{\textit{Qwen3.5 / 05-06 ($n=28$)}} \\
& \textsc{NH} & 56.5 & 61.6 & 62.0 & 60.2 & $+3.7$ \\
& \textsc{FH} & 55.5 & 56.4 & 57.2 & 68.5 & $+13.0$ \\
\addlinespace[3pt]
\multicolumn{7}{@{}l}{\textit{GLM-5.1 / 05-06 ($n=28$)}} \\
& \textsc{NH} & 60.5 & 64.0 & 62.4 & 65.0 & $+4.5$ \\
& \textsc{FH} & 55.0 & 63.0 & 66.0 & 70.5 & $+15.5$ \\
\bottomrule
\end{tabular}

\end{table}

\paragraph{Backbone reproducibility.}
Table~\ref{tab:app_backbone_repro} replicates the headline setting (2026-05-06) under three backbones from different families. The same-direction \textsc{FH}--\textsc{NH} effect survives across all three.

\begin{table}[tbp]
\caption{\textbf{Backbone reproducibility on the headline setting (2026-05-06).} Deepest valid pre-resolution horizon per backbone with explicit scored/launched counts, scored with the \textsc{FutureWorld} scorer of App.~\ref{sec:app_futureworld_scoring}. The within-condition $\Delta$ readouts are in Tab.~\ref{tab:mechanism_readout}; the absolute T$-1$d levels reported here are a complementary view.}
\label{tab:app_backbone_repro}
\centering
\small
\renewcommand{\arraystretch}{1.12}
\setlength{\tabcolsep}{4.0pt}
\begin{tabular}{@{}lccccc@{}}
\toprule
\textbf{Backbone} & \textbf{Deepest valid horizon} & \textbf{\textsc{NH}} & \textbf{\textsc{FH}} & \textbf{\textsc{FH}$-$\textsc{NH}} & \textbf{Scored / launched} \\
\midrule
GPT-5.4              & T$-1$d & 60.8 & 66.9 & $+6.1$  & 28 / 50 \\
Qwen3.5-397B-A17B    & T$-1$d & 60.2 & 68.5 & $+8.3$  & 28 / 50 \\
GLM-5.1              & T$-1$d & 65.0 & 70.5 & $+5.5$  & 28 / 50 \\
\bottomrule
\end{tabular}

\end{table}

\paragraph{Compute-matched check.}
Table~\ref{tab:app_compute_matched} reports the per-question average tool-call count, prompt-token count, and LLM-call count at T-1d for each (backbone, date) cell of the main mechanism table. Numbers are computed from the released \texttt{main\_agent\_stats.json} per checkpoint, averaged over the matched scored set. Across all four cells, \textsc{FH} runs use compute footprints within $\pm 5\%$ of \textsc{NH}, and on Qwen3.5 \textsc{FH} uses fewer tool calls and fewer prompt tokens than \textsc{NH} despite a $+9.3$ point gain in $\Delta_{\textsc{FH}}-\Delta_{\textsc{NH}}$ on the same cell.

\begin{table}[tbp]
\caption{\textbf{Per-question compute footprint at T-1d.} Source: \texttt{main\_agent\_stats.json} per checkpoint, averaged over the matched scored set of each cell. \textsc{FH} runs use comparable tool-call and prompt-token budgets to \textsc{NH} across all four cells, ruling out an extra-compute explanation of the within-condition gain in Tab.~\ref{tab:mechanism_readout}.}
\label{tab:app_compute_matched}
\centering
\small
\renewcommand{\arraystretch}{1.10}
\setlength{\tabcolsep}{5.0pt}
\begin{tabular}{@{}llccc@{}}
\toprule
\textbf{Cell (T-1d)} & \textbf{Cond.} & \textbf{Tool calls} & \textbf{Prompt tok.} & \textbf{LLM calls} \\
\midrule
GPT-5.4 / 05-05 & \textsc{NH} & 22.4 & 159.7\,K & 12.1 \\
                & \textsc{GH} & 19.2 & 135.2\,K &  9.7 \\
                & \textsc{FH} & 23.3 & 156.1\,K & 11.7 \\
\midrule
GPT-5.4 / 05-06 & \textsc{NH} & 36.7 & 262.9\,K & 22.7 \\
                & \textsc{GH} & 33.0 & 233.1\,K & 19.0 \\
                & \textsc{FH} & 34.3 & 266.4\,K & 22.4 \\
\midrule
Qwen3.5 / 05-06 & \textsc{NH} & 29.0 & 407.8\,K & 28.8 \\
                & \textsc{FH} & 25.9 & 369.6\,K & 26.2 \\
\midrule
GLM-5.1 / 05-06 & \textsc{NH} & 32.7 & 287.0\,K & 21.6 \\
                & \textsc{FH} & 34.6 & 311.3\,K & 22.3 \\
\bottomrule
\end{tabular}

\end{table}

\begin{figure}[tbp]
\centering
\includegraphics[width=0.62\textwidth]{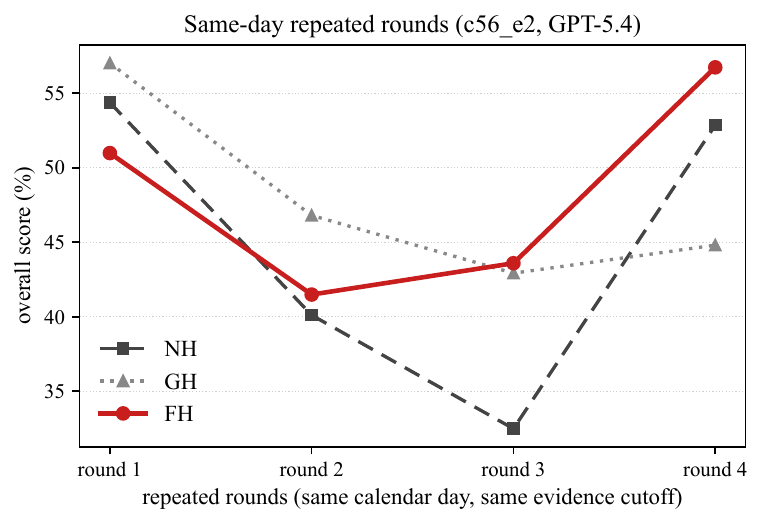}
\caption{\textbf{Same-day repeated-rounds control ($c56\_e2$, GPT-5.4, $n{=}28$).} Four rounds on the same calendar day, same evidence cutoff, scaffold, tools, and budget. \textsc{NH} and \textsc{GH} do not improve when the calendar is held fixed; \textsc{FH} improves by $+5.7$ points, less than half of its calendar-ladder $\Delta$ on the same setting.}
\label{fig:mechanism_sameday}
\end{figure}

\paragraph{Same-day repeated-rounds control.}
Table~\ref{tab:app_supporting_controls} reports the same-day repeated-rounds control referenced in \S\ref{sec:exp_mechanism}. Holding the calendar-time evidence cutoff fixed across four rounds, repetition alone does not improve \textsc{NH} or \textsc{GH}, and \textsc{FH} improves by less than half of its calendar-ladder $\Delta$ on the same setting. Fig.~\ref{fig:mechanism_sameday} visualizes the same data.

\begin{table}[tbp]
\caption{\textbf{Same-day repeated-round control ($c56\_e2$, GPT-5.4, $n{=}28$).} All four rounds use the same calendar-time evidence snapshot. Any change across rounds is attributable to rerunning, stochasticity, or in-arm state evolution rather than calendar-evidence drift. The calendar-ladder $\Delta_{\textsc{FH}}-\Delta_{\textsc{NH}}$ values in Tab.~\ref{tab:mechanism_readout} comfortably exceed the round-to-round dispersion bound observed under \textsc{NH} and \textsc{GH}.}
\label{tab:app_supporting_controls}
\centering
\small
\renewcommand{\arraystretch}{1.12}
\setlength{\tabcolsep}{4.0pt}
\begin{tabular}{@{}lccccc@{}}
\toprule
\textbf{Arm} & \textbf{Round 1} & \textbf{Round 2} & \textbf{Round 3} & \textbf{Round 4} & $\Delta(\mathrm{R4}-\mathrm{R1})$ \\
\midrule
\textsc{NH} & 54.4 & 40.1 & 32.5 & 52.9 & $-1.5$ \\
\textsc{GH} & 57.0 & 46.8 & 42.9 & 44.8 & $-12.2$ \\
\textsc{FH} & 51.0 & 41.5 & 43.6 & 56.7 & $\phantom{-}+5.7$ \\
\bottomrule
\end{tabular}

\end{table}

\subsection{Mechanism audit}
\label{sec:app_futureworld_artifacts}

This section audits the realized writeback artifacts. Its purpose is to verify that the editor behaves like the bounded update mechanism defined in Sec.~\ref{sec:method_feedback}, and that the \textsc{FH}--\textsc{GH} contrast is preserved in the realized artifacts rather than only in the scores.

Table~\ref{tab:app_mechanism_audit} summarizes two complementary surfaces. Panel~A tracks the typed patch stream under \textsc{FH} over the rolling ladders. The first writable checkpoint is add-dominant, as expected when the harness first becomes non-empty, while later checkpoints show a larger number of revise operations, consistent with bounded state refinement rather than monotonic accumulation. Across both cohorts, evidence-handling writes dominate, factor-tracking writes are sparser, and uncertainty-handling writes remain present throughout. This distribution is descriptive rather than normative: the main-text case (\S\ref{sec:exp_case}) reflects the typical $E$+$U$ co-occurrence, while App.~\ref{sec:app_futureworld_cases} adds an $F$+$E$ case to make the factor-tracking axis concrete.

Panel~B audits the \textsc{GH} control at the artifact level. Under the same persistence budget, the generic memory blob is non-empty, but it contains no forbidden typed-harness schema vocabulary. This check matters because the typed-versus-generic contrast would be blurred if the free-form control implicitly reconstructed the same explicit $F/E/U$ interface.

\begin{table}[tbp]
\caption{\textbf{Mechanism audit for the GPT-5.4 rolling ladders.} Panel~A reports the realized typed patch stream under \textsc{FH}. The $F/E/U$ columns count axis assignments of ADD operations, while the final column counts REVISE operations at the same checkpoint. Panel~B audits the matched \textsc{GH} control: the memory blob is non-empty under the same persistence budget, but contains no explicit typed-harness schema vocabulary.}
\label{tab:app_mechanism_audit}
\centering
\small
\renewcommand{\arraystretch}{1.12}
\setlength{\tabcolsep}{5.0pt}

\begin{tabular}{@{}llcccccc@{}}
\toprule
\multicolumn{8}{@{}l}{\textbf{Panel A: Typed harness patch stream (\textsc{FH})}} \\
\midrule
\textbf{Cohort} & \textbf{ckpt} & \textbf{Nonempty qns.} & \textbf{ADD entries} & \textbf{F} & \textbf{E} & \textbf{U} & \textbf{REVISE ops} \\
\midrule
c55 & 2 & 49/50 & 93 & 6 & 72 & 15 & 0 \\
c55 & 3 & 30/50 & 41 & 6 & 18 & 5 & 12 \\
c55 & 4 & 26/44 & 36 & 4 & 8 & 8 & 16 \\
c56 & 2 & 50/50 & 97 & 3 & 76 & 18 & 0 \\
c56 & 3 & 29/50 & 40 & 3 & 13 & 6 & 18 \\
c56 & 4 & 31/47 & 42 & 4 & 7 & 3 & 28 \\
\bottomrule
\end{tabular}

\vspace{0.65em}

\begin{tabular}{@{}lcccc@{}}
\toprule
\multicolumn{5}{@{}l}{\textbf{Panel B: Generic-memory artifact audit (\textsc{GH})}} \\
\midrule
\textbf{Cohort} & \textbf{Deepest ckpt} & \textbf{Nonempty files} & \textbf{Blob size} & \textbf{Typed-schema violations} \\
\midrule
c55 & 4 & 48 & 917.8 bytes & 0 \\
c56 & 4 & 48 & 934.0 bytes & 0 \\
\bottomrule
\end{tabular}

\end{table}

\subsection{Qualitative mechanism cases}
\label{sec:app_futureworld_cases}

This section collects the qualitative mechanism material referenced in the main paper but omitted from the core text for space. Its role is not to introduce a second results section, but to make the typed writeback mechanism legible on concrete unresolved \textsc{FutureWorld} trajectories.

\paragraph{Primary mechanism case (full artifact).}
The case summarized in \S\ref{sec:exp_case} (Hebei province-level inbound migration share, gold $4.48$) covers two of the three axes simultaneously: an $E$-axis routing rule that forces subsequent runs onto the live Baidu Migration UI under the inbound-destination view at the province level, and a $U$-axis cap on confidence when no direct official point is observable. Fig.~\ref{fig:case_study} reproduces the full verbatim writes, the matched-checkpoint controls, and the contrasting \textsc{GH} blob.

\begin{figure}[tbp]
\centering
\fbox{\parbox{0.95\textwidth}{\small
\textbf{Task} (\textsc{FutureWorld} id \texttt{20260430180910705825}; resolved 2026-05-05). Forecast Baidu Migration's province-level inbound migration share for Hebei Province on 2026-05-05, as a percentage value. The benchmark fixes the metric: \emph{province-level inbound destination share for the named province on the named date}, no averaging or substitution. \textbf{Resolved gold: 4.48.}
\\[0.4ex]
\textbf{Typed schema.} The harness has three axes: $F$ \emph{(what to monitor)}, $E$ \emph{(where to look and how to verify)}, $U$ \emph{(when to cap confidence)}. \textsc{NH} writes nothing; \textsc{GH} writes a free-form scratchpad of equal byte and write-call budget; \textsc{FH} writes typed entries.
\\[0.4ex]
\textbf{\textsc{FH} ck1 (T-4d), empty harness.} Prediction $\mathbf{35.2}$. The agent searches old Baidu JSONP endpoints, third-party ranking pages, and historical migration commentary, anchoring on a wrong-surface metric and overshooting by $\sim$31 pp.
\\[0.4ex]
\textbf{Editor writes (verbatim from \texttt{harness/.../legacy\_versions/ck\{2,3\}/SKILL.md}).}\\
$\bullet$ Between ck1 and ck2 ($U1$, axis: Uncertainty Handling). \emph{``When the exact province-level inbound-share point for the target or close holiday analog dates cannot be directly observed on the official surface, and the estimate relies on holiday-flow analogies or traffic forecasts. Cap confidence at low-to-medium and state the flip condition as obtaining a direct official province-level point or a close same-metric historical analog; do not commit firmly from general return-flow narratives alone.''}\\
$\bullet$ Between ck2 and ck3 ($E1$, axis: Evidence Handling). \emph{``When estimating this question's Hebei move\_in province-level share, and web search/old JSONP endpoints return forecast articles, city rankings, or 404s instead of the exact metric. Route first to the live Baidu Migration front end, switch explicitly to the inbound-destination view at the province level, and use only that surface as the direct anchor. Treat guessed legacy JSONP/API paths and third-party ranking pages as non-authoritative support unless they reproduce the same province-level metric.''}
\\[0.4ex]
\textbf{\textsc{FH} after the writes.} ck2 (T-3d, $U1$ active) predicts $5.46$; ck3 (T-2d, $U1{+}E1$ active) predicts $5.23$; ck4 (T-1d) predicts $\mathbf{5.24}$. Once $E1$ is in the harness, every subsequent run routes to the live Baidu Migration UI under the inbound-destination view at the province level as the direct anchor, treating legacy JSONP and third-party rankings as background.
\\[0.4ex]
\textbf{Matched-checkpoint controls and reference} (T-1d, same task, same scaffold/tools/budget):
\begin{center}
\begin{tabular}{@{}lcccc@{}}
\toprule
& \textsc{FH} & \textsc{NH} & \textsc{GH} & gold \\
\midrule
prediction          & $\mathbf{5.24}$ & $27.18$ & $27.30$ & $4.48$ \\
$|\text{pred}-\text{gold}|$ & $\mathbf{0.76}$ & $22.70$ & $22.82$ & --- \\
\textsc{FutureWorld} score & $\mathbf{0.85}$ & $0.00$ & $0.00$ & --- \\
\bottomrule
\end{tabular}
\end{center}
\textbf{What \textsc{GH} wrote (verbatim, ck4 free-form memory).} \emph{``\dots Treat the target as Hebei's top inbound source-province share only if the wording still supports that reading\dots Keep using Baidu Huiyan's undocumented JSONP endpoints, especially \texttt{lastdate.jsonp} and \texttt{provincerank.jsonp}\dots''}
}}
\caption{\textbf{Case study (full artifact): typed writeback encodes a metric-routing rule, not the answer.} The benchmark question fixes a specific metric (province-level inbound destination share); ranking-page and JSONP surfaces expose a different number ($\sim$27) that is a wrong-surface readout. \textsc{FH} ck1 makes the same wrong-surface error (35.2). The next two editor writes do not record an answer: $U1$ caps confidence under uncertainty, and $E1$ writes the procedural rule \emph{which surface to read first}. Subsequent \textsc{FH} runs execute that rule and converge to within one point of gold, while matched-budget \textsc{GH} writes a memory blob that locks in the \emph{opposite} interpretation and stays $\sim$23 points off; \textsc{NH} drifts between near-zero and $\sim$27 across runs. The persistence of the metric-definition discovery across runs---rather than its rediscovery from scratch each checkpoint---is what the typed harness contributes. Artifacts: \texttt{log/FutureWorld/milkyway\_v5/c55/\{nh,gh,fh\}/20260430180910705825/}.}
\label{fig:case_study}
\end{figure}

We add one secondary case below to demonstrate a different write pattern in which $F$ (factor monitoring) and $E$ (forward-page routing) act jointly.

\paragraph{Secondary case: redirecting evidence for Xinxiang AQI ($F$+$E$).}
\textsc{FutureWorld} id \texttt{20260430180910740621} asks for the Xinxiang, Henan city-level AQI on 2026-05-05; the resolved gold value is $91$. At checkpoints 1--4, \textsc{FH} predicts $43,37,58,58$ while repeatedly searching historical AQI tables and city-ranking pages that do not expose the target-day value reliably. Between checkpoints~4 and~5, the editor writes two typed entries. The $F$ entry says to track local current AQI, the 24--48h trajectory, and same-day dispersion or ozone weather before using neighboring-city context as a tie-breaker; the $E$ entry says that, when historical daily AQI pages are inaccessible and the target day is within roughly 24 hours, the next run should first check whether a city-level next-day AQI forecast page exists, then use live trackers mainly for plausibility and offset checks. Neither entry writes the answer. At checkpoint~5, the next \textsc{FH} run follows this rule, routes to a date-matched QWeather Xinxiang forecast page citing the China National Environmental Monitoring Centre, observes a forward AQI of $88$, and uses NMC weather plus nearby-city AQI only as plausibility checks. Its final prediction is $88$, three points from gold. At the same matched checkpoint, \textsc{NH} predicts $78$ and \textsc{GH} predicts $64$. Artifacts at \texttt{log/FutureWorld/milkyway\_v5/c55/\{nh,gh,fh\}/20260430180910740621/} and \texttt{.../fh/harness/20260430180910740621/SKILL.md}.

\section{Runtime interfaces and prompt excerpts}
\label{sec:app_prompt_case}

This appendix shows only the load-bearing runtime clauses: the shared prior used by the BaseAgent, the editor interface, and the structured editor output schema. It is not meant to restate the method semantics from Sec.~\ref{sec:method_overview} and Sec.~\ref{sec:method_feedback}. Concrete artifacts from real \textsc{FutureWorld} runs appear in App.~\ref{sec:app_futureworld_cases}.

\subsection{BaseAgent prompt excerpt}
\label{sec:app_prompt_worker}

The BaseAgent prompt defines the shared prior $B$. It governs question audit, evidence gathering, temporal discipline, and structured finalization; it does not contain question-local harness content. We show only the clauses that are load-bearing for the method description in Sec.~\ref{sec:method_feedback}.

\begin{tcblisting}{guiguardprompt,title=BaseAgent prompt (load-bearing clauses),listing only,listing options={basicstyle=\ttfamily\small,breaklines=true,columns=fullflexible}}
You are a research-driven future prediction agent. The event is unresolved at checkpoint time. Your job is to gather predictive evidence and produce a well-grounded prediction.

Rules:
- Audit the question: target, answer space, constraints, and resolution rule.
- Anchor on a primary source and trace what has changed.
- Before each tool call, state what missing evidence it should resolve.
- After each result, separate evidence, interpretation, and guess.
- If the same artifact recurs without new signal, switch source or finalize with an explicit gap.
- End with the structured finalization action that records the prediction, supporting evidence, counterevidence, and unresolved concerns.
- Keep confidence bounded when evidence is thin.
\end{tcblisting}

In the experiments, the structured finalization payload is the source of the stored prediction and checkpoint note; the method does not rely on any specific prose formatting outside that payload.

\subsection{Harness editor prompt and output schema}
\label{sec:app_prompt_editor}

The Harness Editor does not answer the prediction question. Instead, it reads the ordered checkpoint note history for the same unresolved question, diagnoses temporal contrasts, and proposes a bounded harness update.

\begin{tcblisting}{guiguardprompt,title=Harness editor prompt (load-bearing clauses),listing only,listing options={basicstyle=\ttfamily\small,breaklines=true,columns=fullflexible}}
You are the Harness Editor. Do not answer the prediction question. Read the ordered checkpoint note history for the SAME unresolved question, diagnose temporal contrasts, and propose a bounded harness update to the future prediction harness.

Rules:
- Prefer a null update over a weak update.
- Write only reusable procedural guidance; do not write predicted answers, one-off statuses, raw evidence, or checkpoint-specific facts.
- Axes: F (factor tracking), E (evidence handling), U (uncertainty managing). Route source or verification lessons to E by default.
- If an active entry already covers the lesson, revise it; if contradicted or subsumed, deprecate it; otherwise add it.
- Call `propose_harness_update` exactly once with both `signal` and `update`.
\end{tcblisting}

The editor is constrained to a structured output that separates the pre-resolution signal from the bounded harness update:

\begin{tcblisting}{guiguardprompt,title=Editor output schema,listing only,listing options={basicstyle=\ttfamily\small,breaklines=true,columns=fullflexible}}
{
  "signal": {
    "prediction_divergence": [...],
    "diagnostic_findings":   [...]
  },
  "update": {
    "add":       [{"axis": "F|E|U", "when": "...", "guidance": "..."}],
    "revise":    [{"id": "...",     "when": "...", "guidance": "..."}],
    "deprecate": [{"id": "...",     "reason": "..."}]
  }
}
\end{tcblisting}

This schema makes Eqs.~\ref{eq:method_signal}--\ref{eq:method_apply} concrete: \texttt{signal} corresponds to $\Psi(H_t,N_{\le t})$, \texttt{update} corresponds to $\mathrm{Edit}(H_t,I_t)$, and $\mathrm{Apply}$ is the deterministic runtime step of App.~\ref{sec:app_runtime}.

\end{document}